\documentclass[journal,twoside,web]{ieeecolor}
\usepackage{adjustbox}
\usepackage{algorithmic}
\usepackage{amsfonts}
\usepackage{amsmath}
\usepackage{amssymb}
\usepackage{array}
\usepackage{arydshln}
\usepackage{authblk}
\usepackage{booktabs}
\usepackage{caption}
\usepackage{cite}
\usepackage{comment}
\usepackage{diagbox}
\usepackage{graphicx}
\usepackage{hyperref}
\usepackage{makecell}
\usepackage{multicol}
\usepackage{multirow}
\usepackage{pifont}
\usepackage{tabularx}
\usepackage{tabulary}
\usepackage{textcomp}
\usepackage{tmi}
\usepackage[font=footnotesize,labelfont=bf]{caption}
\usepackage{xcolor}%

\def\BibTeX{{\rm B\kern-.05em{\sc i\kern-.025em b}\kern-.08em
    T\kern-.1667em\lower.7ex\hbox{E}\kern-.125emX}}
    
\begin{document}

\venue{Published in Pattern Recognition Letters, Volume 190, Pages 73-80, April 2025}

\title{Surgical Text-to-Image Generation}

\author[1,4]{Chinedu Innocent Nwoye}
\author[1]{Rupak Bose}
\author[1]{Kareem Elgohary}
\author[1]{Lorenzo Arboit}
\author[2,4]{Giorgio Carlino}
\author[3,4]{Joël L. Lavanchy}
\author[2,4]{Pietro Mascagni}
\author[1,4]{Nicolas Padoy}

\affil[1]{University of Strasbourg, CNRS, INSERM, ICube, UMR7357, Strasbourg, France}
\affil[2]{Fondazione Policlinico Universitario Agostino Gemelli IRCCS, Rome, Italy}
\affil[3]{University of Basel, Switzerland}
\affil[4]{IHU Strasbourg, France}
\affil[ ]{ }
\affil[ ]{\textbf{\emph{Project page}}: \url{https://camma-public.github.io/endogen}}

\renewcommand{\Authfont}{\bfseries}
\renewcommand{\Affilfont}{\mdseries\small \itshape}

\twocolumn[{%
\renewcommand\twocolumn[1][]{#1}%
\maketitle
\unskip
\unskip
\bigskip
\begin{center}
    \centering
    \captionsetup{type=figure}
    \includegraphics[width=.999\textwidth,height=6.02cm]{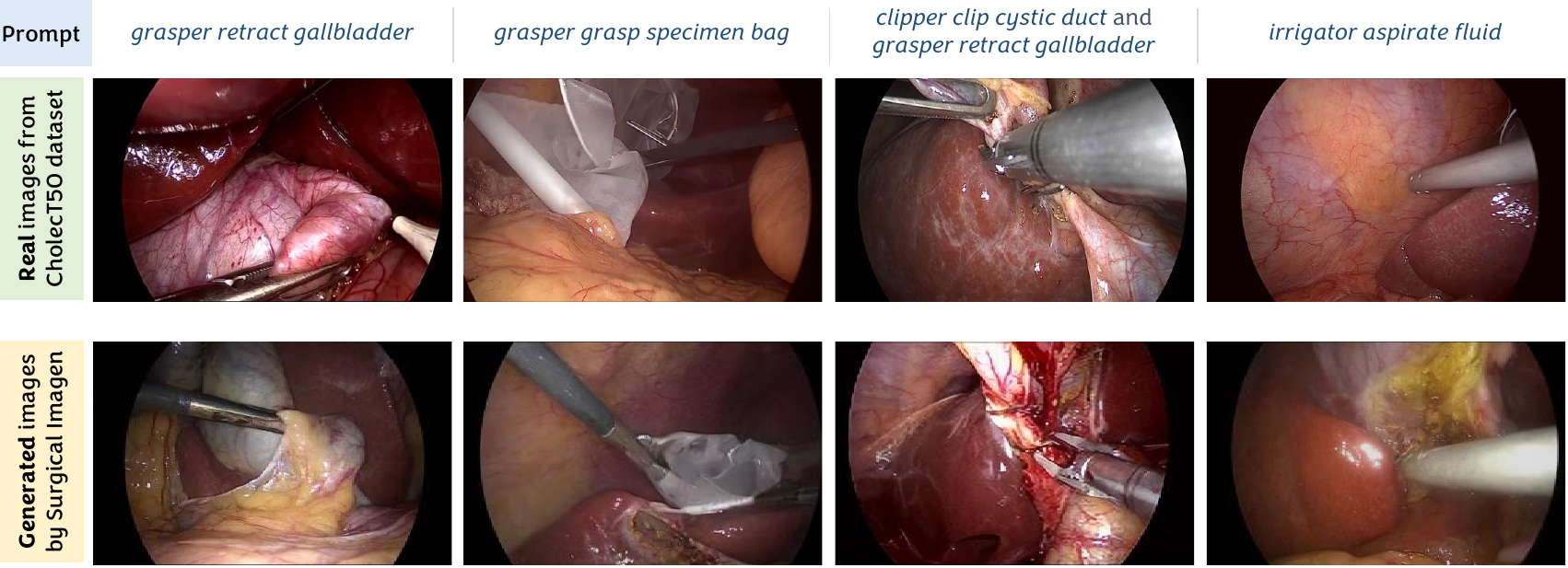}
    \captionof{figure}{Qualitative comparison between real surgical images from the CholecT50 dataset \cite{nwoye2022rendezvous} and synthetic images generated by Surgical Imagen. The model inputs are only triplet-based textual prompts in the form of ({\it instrument verb target}) syntax. 
    }
\end{center}%
\medskip
}]

\begin{abstract}
Acquiring surgical data for research and development is significantly hindered by high annotation costs and practical and ethical constraints. Utilizing synthetically generated images could offer a valuable alternative.
In this work, we explore adapting text-to-image generative models for the surgical domain using the CholecT50 dataset, which provides surgical images annotated with action triplets (instrument, verb, target). 
We investigate several language models and find T5 to offer more distinct features for differentiating surgical actions on triplet-based textual inputs, and showcasing stronger alignment between long and triplet-based captions. 
To address challenges in training text-to-image models solely on triplet-based captions without additional inputs and supervisory signals, we discover that triplet text embeddings are instrument-centric in the latent space. Leveraging this insight, we design an instrument-based class balancing technique to counteract data imbalance and skewness, improving training convergence. 
Extending Imagen, a diffusion-based generative model, we develop {Surgical Imagen} to generate photorealistic and activity-aligned surgical images from triplet-based textual prompts. 
We assess the model on quality, alignment, reasoning, and knowledge, achieving FID and CLIP scores of 3.7 and 26.8\% respectively. Human expert survey shows that participants were highly challenged by the realistic characteristics of the generated samples, demonstrating Surgical Imagen's effectiveness as a practical alternative to real data collection.


\end{abstract}



\vspace{-2.5em}
\section{Introduction}
\label{sec:introduction}
The exploitation of surgical data for AI development is crucially limited by high costs of data annotation, privacy concerns, and ethical and practical constraints \cite{nwoye2021deep}. Despite the promise of advanced recording technologies \cite{shah2020surgical} in the operating room (OR), access to comprehensive surgical datasets remains restricted, impeding the development and training of effective AI models for surgical assistance and education \cite{vercauteren2019cai4cai,padoy2019machine}. Available datasets often lack critical cases, particularly those illustrating surgical complications \cite{stam2022prediction}. There is also a significant imbalance in the representation of surgical phases \cite{stam2022prediction}, with the most critical ones, such as clipping and cutting in laparoscopic cholecystectomy for instance, being notably brief. The under-representation of these essential yet short workflow phases in surgical videos impacts the AI model's ability to effectively learn and assist in the most crucial scenarios. Moreover, the manual annotation of surgical data is labor-intensive and depends on the availability of expert surgical professionals, introducing further uncertainties into the training process. This highlights the urgent need for innovative solutions to address these data challenges.

In response, we propose \textit{Surgical Imagen}, a diffusion-based text-to-image generative model designed to synthesize photorealistic surgical images from triplet-based textual prompts. Building on the capabilities of the Imagen framework \cite{saharia2022photorealistic}, which integrates a large language model, a diffusion model, and a super-resolution component, Surgical Imagen aims to generate high-fidelity surgical images without complex input requirements.
Unlike existing approaches that may lack photorealism or necessitate intricate inputs such as reference images \cite{pfeiffer2019generating,sharan2021mutually,colleoni2021robotic,kaleta2023lc}, segmentation masks \cite{marzullo2021towards,mazzocchetti2024semantic,allmendinger2023navigating}, or instrument position priors \cite{allmendinger2023navigating}, which are often difficult to obtain, our model relies solely on textual inputs to achieve realistic surgical image generation.

Our approach leverages the CholecT50 dataset \cite{nwoye2022rendezvous}, which provides surgical images annotated with triplet-based labels (instrument, verb, target). These triplet-based prompts, such as "\textit{clipper clip cystic duct}" or "\textit{hook dissect liver}," capture the essential semantics of surgical tool-tissue interactions in a concise format \cite{nwoye2020recognition}. Initial experiments reveal that these triplets are semantically aligned with longer descriptive captions in the text embedding space, validating their use for generating meaningful and contextually accurate images.

A significant challenge in utilizing the CholecT50 dataset is the high class imbalance. Our analysis shows that triplet text embeddings cluster semantically around their instrument categories, which lead us to develop an instrument-based class balancing technique. This method enhances model training by ensuring more effective convergence when conditioned solely on textual prompts.

To validate the effectiveness of Surgical Imagen, we employ both human expert evaluations and a comprehensive set of automated metrics, including FID and CLIP scores \cite{heusel2017gans,betzalel2022study,parmar2022aliased}. These evaluations assess various aspects of the generated images, such as quality, alignment with textual prompts, reasoning, knowledge, and robustness. 
Our results demonstrate that Surgical Imagen, with an FID score of 3.7, generates more realistic surgical images than the baseline and effectively aligns the generated image contents with the text prompts, as reflected by a CLIP score of 26.8.

Surgical Imagen offers a transformative solution for generating high-quality surgical images, addressing the critical shortage of comprehensive surgical datasets. 
By filling gaps in existing datasets and providing a robust methodology for surgical image generation, Surgical Imagen simplifies the complexities of dataset requirements for the development of large-scale interventional AI.
The code will be made available to encourage further research and collaboration in this vital area of surgical AI.
\section{Related Work}
\label{sec:literature}

\subsection{Image Generation Advancements}
The field of image generation has undergone significant advancements in recent years, largely driven by seminal works that have introduced innovative methodologies. One such milestone is the introduction of Generative Adversarial Network (GAN) \cite{goodfellow2014generative}, which laid the groundwork for estimating deep generative models by adversarial training.
This approach revolutionized the field by enabling models to generate images that closely resemble real-world data through a competitive process between a generator and a discriminator.
Subsequently, diffusion processes \cite{ho2020denoising} that can generate high-quality images by iteratively refining noise-corrupted versions is showcased. This method not only produces visually pleasing results but also offers robustness against noise and artifacts, thereby improving the overall quality of generated images.
Building upon this foundation, \cite{rombach2022highresolution} incorporate latent space training to diffusion based denoising to capture complex image features and generate high-resolution images with fine-grained details and high fidelity.

\subsection{Text-to-Image Synthesis}

The evolution from general image generation to text-to-image synthesis has expanded the capabilities of generative models. Early efforts, as seen in \cite{reed2016generative}, demonstrated the feasibility of generating images from textual descriptions, albeit with limitations in resolution. 
A pivotal advancement in text-to-image synthesis was the incorporation of attention mechanisms \cite{xu2017attngan}, allowing models to focus on specific words during image generation. This innovation enhanced the realism of generated images by capturing finer details, addressing limitations in earlier models.
Recent research has shifted towards diffusion-based generative models, surpassing the quality of traditional GANs in specific scenarios \cite{karras2022elucidating,ramesh2022hierarchical,saharia2022photorealistic,dhariwal2021diffusion}. Advancements such as vector quantization, use of codebook, are incorporated in the diffusion models \cite{gu2022vector}, enabling the handling of more complex scenes in text-to-image generation.

\subsection{Surgical Data Science}

In the domain of surgical data science, the synthesis of realistic surgical images holds immense promise for training and augmenting datasets. Previous research predominantly relies on image-to-image translation, utilizing various input sources such as reference images, computer simulations, and segmentation maps \cite{pfeiffer2019generating,sharan2021mutually,marzullo2021towards,colleoni2021robotic}. Initial efforts in generating synthetic surgical images using simple GAN models were demonstrated by \cite{8396171}. Recent methodologies have incorporated text prompts as conditioning inputs but primarily focus on grayscale images in medical imaging domains such as lung X-rays and magnetic resonance images \cite{chambon2022adapting,pinaya2022brain}. Notably, \cite{allmendinger2023navigating} demonstrates the generation of surgical RGB images from text prompts, albeit in conjunction with position priors, such as segmentation masks of instruments and anatomies, as input prompts.

Our method aims to push the boundaries of text-to-image synthesis by relying solely on textual inputs for its reduced resource requirements. By leveraging diffusion models, attention mechanisms, and codebook-based approaches, our method integrates large language models to address the unique challenges of generating surgical data, providing a novel solution for generating realistic and diverse surgical images.

\section{Method}
\label{sec:method}
\begin{figure*}
    \centering
    \includegraphics[width=0.8\textwidth]{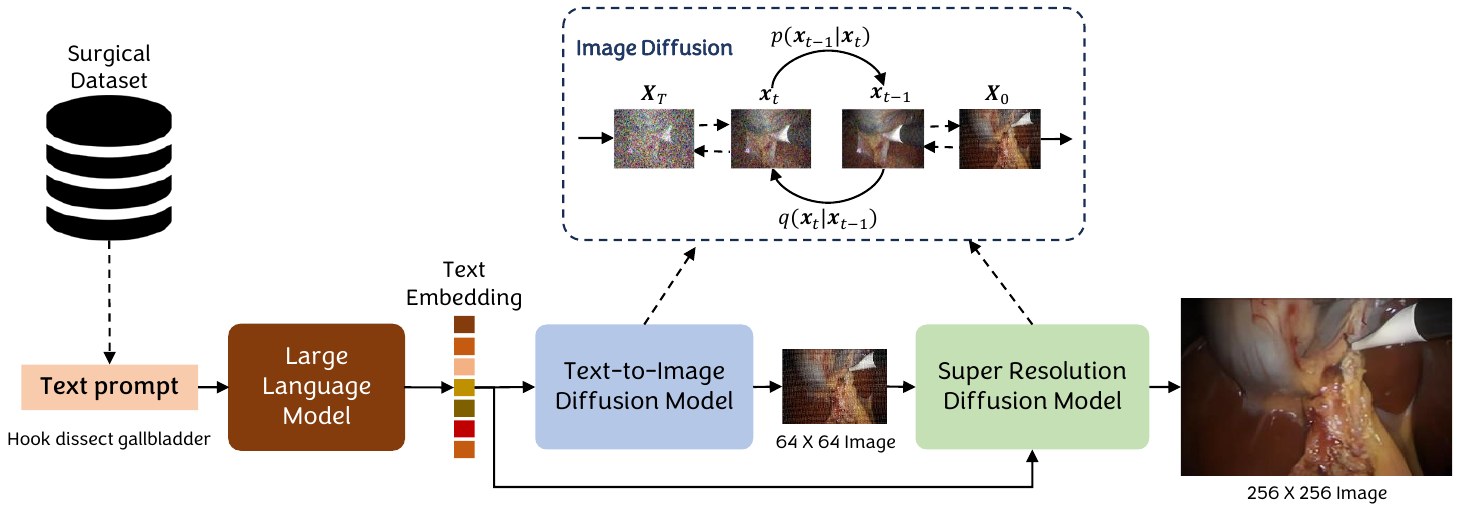}
    \caption{Surgical Imagen: The architecture uses a frozen text encoder to encode the input text into text embeddings. A conditional diffusion model maps the text embedding into a low resolution image, afterward upsamples it to the desired higher resolution via a text-conditional super-resolution diffusion model.}
    \label{fig:model}
\end{figure*}
Built upon the robust Imagen framework \cite{saharia2022photorealistic}, our method, Surgical Imagen, is conceptually divided into (1) a large language model for text encoding and text conditioning, (2) a text-to-image diffusion model for image generation, and (3) a super-resolution diffusion model for upsampling the image to a desired resolution, as shown in Fig. \ref{fig:model}.
In this section, we provide a detailed overview of Surgical Imagen, including our investigation into the suitability and utilization of triplet-based captions for text conditioning. Additionally, we introduce the instrument-based triplet class balancing technique employed to enhance the convergence of our triplet-conditioned diffusion model. We also discuss the super-resolution components and the training process of the proposed model.

\subsection{Surgical Dataset}
Surgical Imagen is tailored for text-to-image generation in laparoscopic cholecystectomy surgery. 
It is trained to generate the surgical images from triplet-based textual prompts.   
These prompts, known as surgical action triplets \cite{nwoye2020recognition}, encapsulate surgical activities in the format \textlangle instrument verb target\textrangle, representing tool-tissue interactions in endoscopic videos.
We utilize CholectT50 \cite{nwoye2022rendezvous} - a dataset for surgical action recognition comprising frames sampled from a set of laparoscopic cholecystectomy surgical videos and the corresponding triplet annotations. 
Given the dataset $D=\{X,T\}_{i=1}^n$ with $n$ surgical images where $X_i$ is the $i^{th}$ image, $T_i$ is the corresponding set of triplet labels:
$\{ \textlangle\text{\it instrument},\text{\it verb},\text{\it target}\textrangle _1,\cdots,\textlangle\text{\it instrument},\text{\it verb},\text{\it target}\textrangle _k \}$ and $k$ is the number of triplet combinations present in the respective frame. 
Our objective is to generate an image $\hat{X}$ such that its content aligns with the tool-tissue interactions depicted by the triplet labels, $\hat{X} \sim p(X_i | T_i)$.
We rely on a language model to extract meaningful embeddings from the triplet captions for text-conditioning of our generative model.

\begin{figure}[h!]
    \centering
    \includegraphics[width=0.9\linewidth]{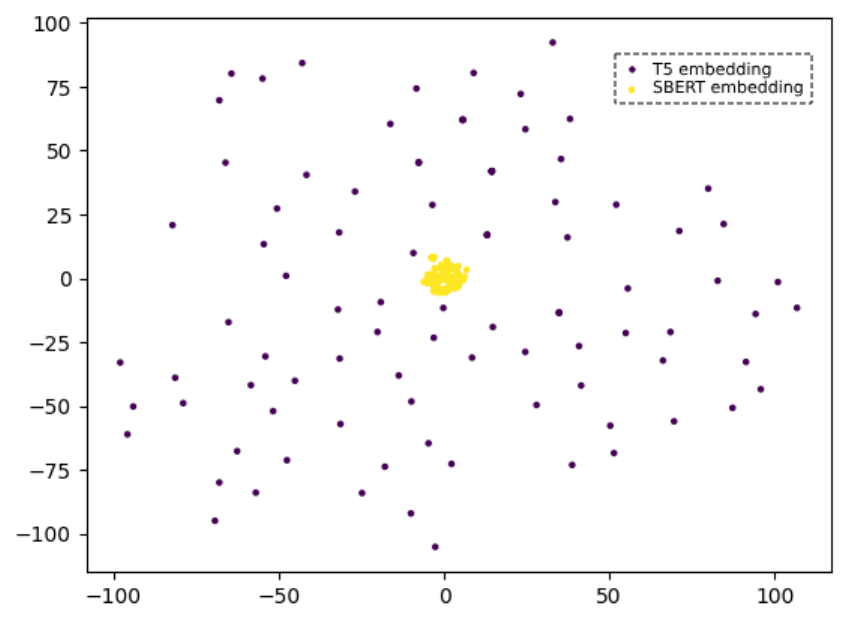}
    \caption{Visualization of the combined latent space of the first two principal components for SBERT (yellow) and T5 encoder. The T5 latent vectors exhibit greater separation compared to SBERT, indicating a more diverse sampling strategy for capturing subtle semantic information in surgical language.}
    \label{fig:TSNE-sbert-t5}
\end{figure}

\begin{figure}[h!]
    \centering
    \includegraphics[width=0.90\linewidth]{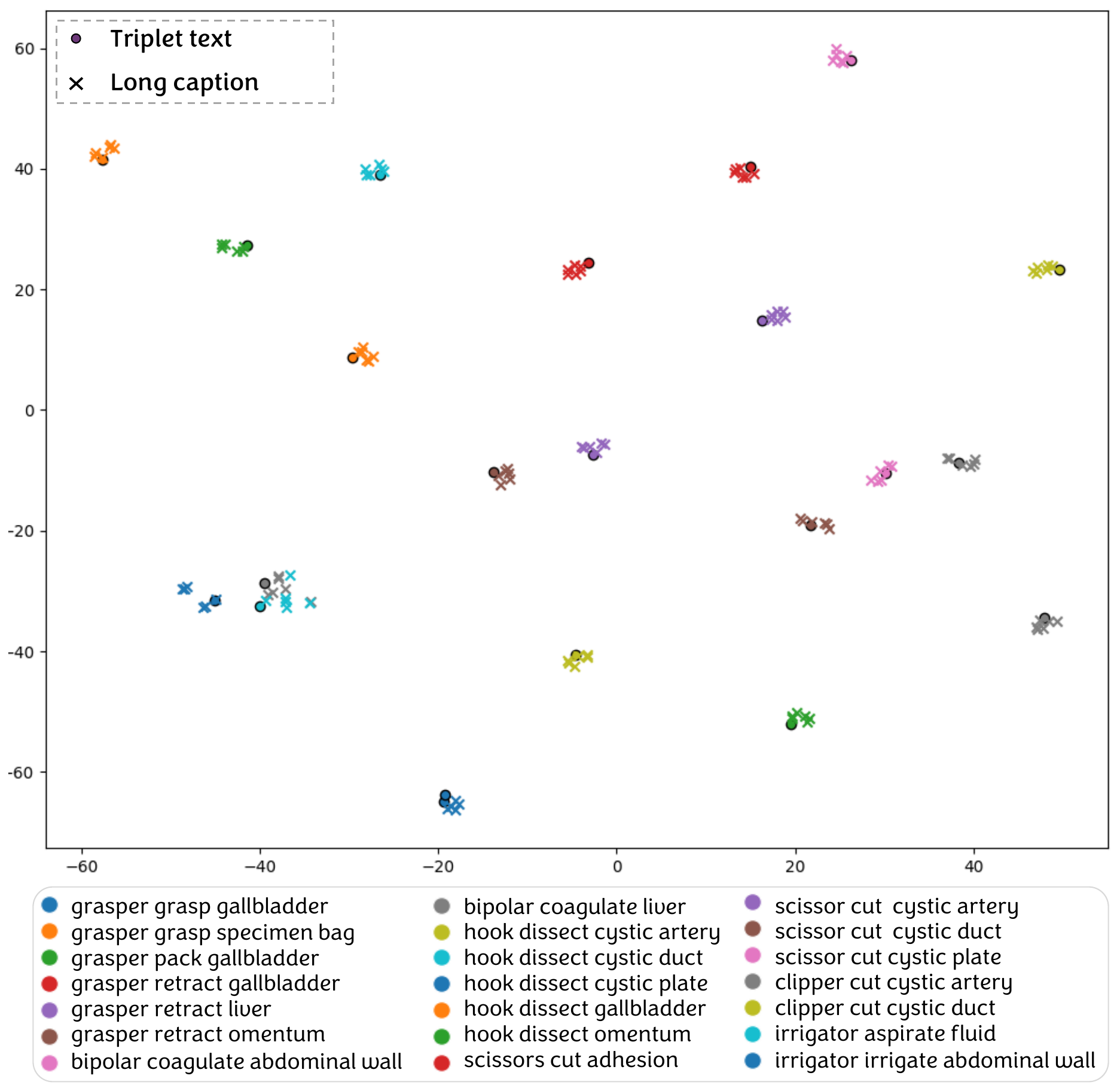}
    \caption{Visualization of the alignment between triplet and longer caption embeddings, showcasing the capability of triplet-based captions to capture fundamental surgical semantics in the longer captions.}
    \label{fig:TSNE-triplet-caption-alignment}
\end{figure}

\begin{figure*}[!th]
    \centering
    \includegraphics[width=0.9\linewidth]{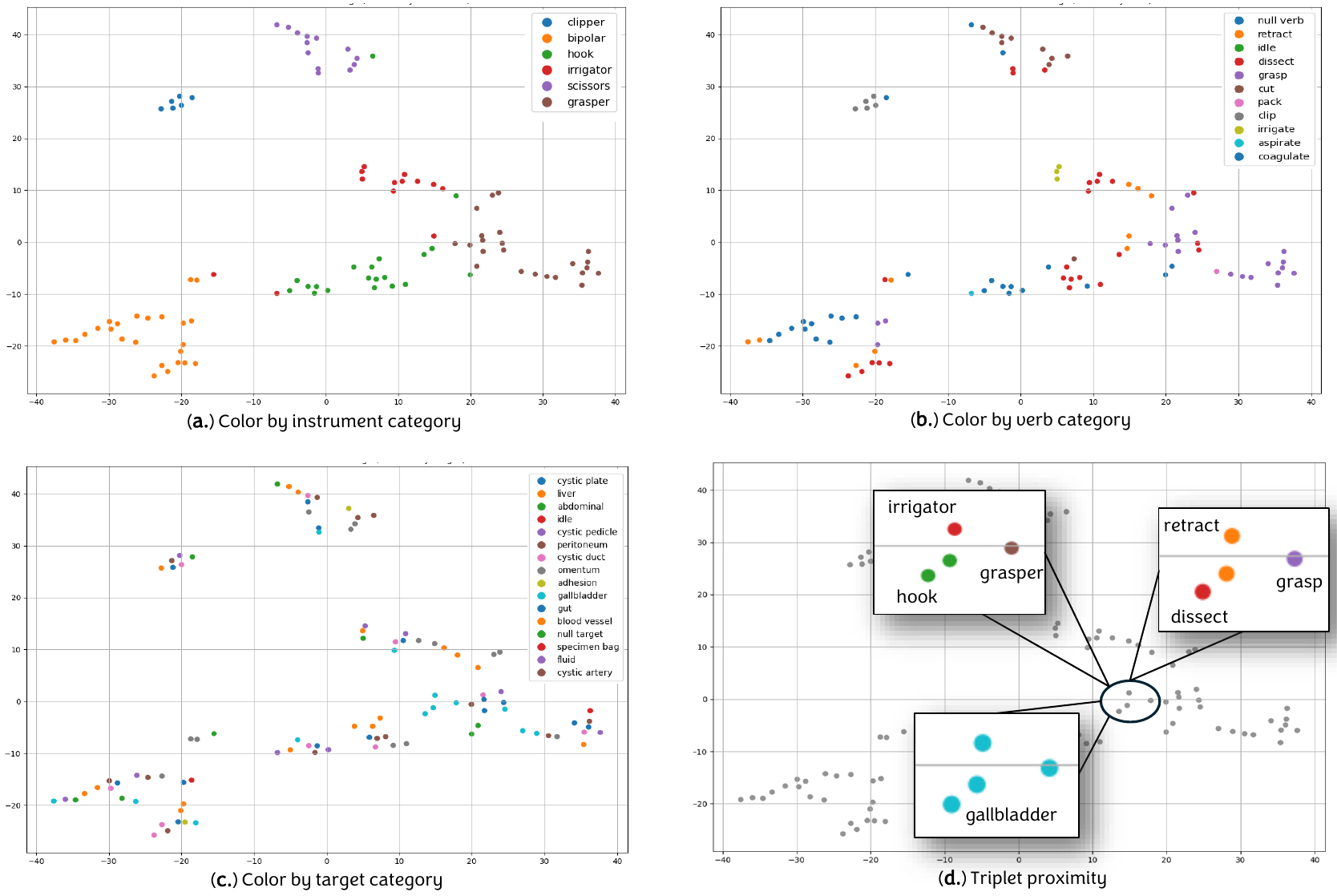}
    \caption{Visualization of disparity and proximity in surgical text embeddings, illustrating the root word clusters in triplet-base surgical language.}
    \label{fig:TSNE-classbal}
\end{figure*}

\subsection{Surgical Language Modeling}

\subsubsection{Choice of Text Encoder}
Text encoders are pivotal in converting textual input sequences $T$ into a sequence of hidden contextualized embeddings $E$, laying the groundwork for downstream tasks in natural language processing (NLP) and text-to-image generation. Our initial investigation compares a decoder-only language model, Sentence-BERT (SBERT) \cite{reimers2019sentence}, with an encoder-decoder counterpart, Text-to-Text Transfer Transformer (T5) \cite{raffel2020exploring}. As depicted in Fig. \ref{fig:TSNE-sbert-t5}, the T5 embeddings exhibit greater separation in the latent space compared to Sentence-BERT embeddings. This indicates a more diverse sampling strategy, crucial for capturing detailed semantic information in surgical language. For instance, while Sentence-BERT may overlook subtle differences between highly co-related surgical actions, such as \textit{"hook dissect cystic artery"} and \textit{"hook dissect cystic plate"}, T5's robust separation capabilities better highlight these distinctions. Encoder-decoder models are optimized for sequence-to-sequence tasks by minimizing entropy for specific objectives, making them better suited for tasks where task probabilities, rather than similarity-based metrics, are paramount. Following these findings, we choose T5 for text encoding in our model.

\subsubsection{Triplet-Caption Semantic Alignment}

We investigate the alignment between triplet-based captions and longer, more descriptive subtitles in the latent space to assess the fidelity of triplet-based captions in capturing surgical activity semantics. 
Here, we select the most frequent 21 triplet categories in the CholecT50 dataset. Given a triplet category text $\hat{T}$ refers to as a short subtitle, we generate $N=6$ varied and longer captions $\overline{T}$ depicting the likely free form captions of the activity represented by the triplet (e.g. \textit{"hook dissect cystic plate."} to \textit{"The hook is applied to dissect the cystic plate, ensuring its safe separation from adjacent tissues."}). The long captions were validated by domain experts to mitigate inherent bias in AI generated text labels. We obtain the text embeddings $E$ for $\hat{T}$ and $\overline{T}$ and compute their angular alignment distributions. Our analysis reveals a mean cosine similarity score of $0.86\pm0.011$, indicating strong alignment between triplet-summarized and longer captions.
Figure \ref{fig:TSNE-triplet-caption-alignment}  visually demonstrates this alignment, highlighting the capability of triplet-based captions to approximate the semantics of longer captions accurately. 
This alignment reinforces the advantage of using triplet-based captions for quick prompting during model inference.

\subsubsection{Class Balancing Text Conditioning}
In text-to-image generation tasks, the text embeddings serve as crucial inputs alongside noise in the diffusion model. During training, the diffusion process denoises the image to recover a clean version, with an effective model prioritizing the influence of text embeddings over noise, particularly relevant during inference where pure noise is utilized. However, in datasets like CholecT50, characterized by class imbalances, the denoising process may become skewed towards the most frequent classes, potentially hindering the model's ability to fully leverage the semantic information embedded in the text. Despite efforts to address this imbalance through class-balancing approaches based on triplet categories, meaningful improvements have been elusive. This challenge may be attributed to the varying contributions of subwords within the triplets, leading to inconsistent oversampling or undersampling across classes.

Acknowledging the limitations of relying solely on triplet-based text, previous efforts have sought to augment triplet-based captions with additional information such as phase details, segmentation masks, tool position priors, reference images, or even conversion to longer captions using caption expansion models. However, achieving optimal model performance by solely using textual inputs remains a significant concern.


Our study investigates the alignment of language model embeddings across triplet classes to identify the most influential subwords for tokenization. Figure~\ref{fig:TSNE-classbal} visualizes the text embeddings of 100 unique triplets from the CholecT50 dataset, with each point representing a triplet’s position in the embedding space. In Figures~\ref{fig:TSNE-classbal}(a-c), we measure embedding distances for each triplet and align them based on subword embeddings to identify the primary contributors to embedding proximity. Analysis in Figure~\ref{fig:TSNE-classbal}(a) highlights the dominance of instrument-related subwords over verbs (Fig.\ref{fig:TSNE-classbal}(b)) and targets (Fig.\ref{fig:TSNE-classbal}(c)). The zoomed-in views in Figure~\ref{fig:TSNE-classbal}(d) highlight regions where triplets with common root words—such as instrument, action, and anatomy—tend to cluster together. The embeddings of triplets with common subwords or semantically related actions, such as \textit{retract} versus \textit{grasp}, tends to cluster together in latent space, underscoring the semantic structure learned by the model. These zoomed-in views offer deeper insights into how the model captures semantic similarities and influences the organization of the latent space, revealing which subwords mostly drive the proximity of triplets.

Based on these insights, we propose an instrument-based class balancing approach to enhance convergence and allow the model to effectively learn the underrepresented triplet classes. This method involves oversampling less frequent triplets based on the inverse frequency of their associated instruments, improving model training effectiveness and handling imbalanced text conditioning in text-to-image generation tasks.

\subsection{Text-to-Image Generation}
Although the image encoder generates useful representation of the triplet-based captions, a method is needed to generate an image leveraging this representation.
We achieve this using a diffusion model which is trained by forward and reverse diffusion process \cite{ho2020denoising}.
The forward diffusion iteratively corrupts the training image $x_0$ through the addition of noise $\hat{n} \sim N(0,I)$ to obtained a noisy image $x_t$ at different time steps $t$. The image corruption happens in a controlled fashion such that the variance does not explode.
The reverse diffusion learns to recover the target image $x_0$ by predicting the added noise $\epsilon_t$ at each timestep $t$ which is in turn subtracted from $x_t$ to obtain a cleaner image $x_{t-1}$ until the target image $x_0$ is achieved.
Meanwhile, during inference, the input image $x_t \sim N(0,I)$ is given as a pure noise without any forward diffusion process.

The diffusion model employs an image encoder to generate an image embedding $e_I$ from the input noisy image $x_t$.
The image encoder and its decoder are based on Attention U-Net \cite{nichol2021improved}.
Given that the image corruption happens over a time sequence, the diffusion model utilizes a timestep encoder to inject timestep information into the denoising step. The timestep encoder is implemented using positional encoding where a unique timestep embedding $e_S$ is generated for each timestep and used in conditioning the model.
To incorporate semantic information about the image, text embeddings $e_T$ from the T5 text encoder are pooled and added to the timestep $e_S$ and image feature $e_I$ embeddings by cross attention at several resolutions for text-conditioning.
The combined embedding sequence ($e_I, e_S, e_T$) is fed into the image decoder part of the Attention U-Net to learn to predict the total noise $\epsilon_t$ present in the noisy image at the given timestep $t$.

\subsection{Super-Resolution}
The base text-to-image generator outputs a small $64 \times 64$ image. Surgical Imagen employs a super-resolution model to upsample the image to a desired size, typically $256 \times 256$ in this study.
Our first exploration of unconditional super-resolution failed to produce high-resolution details and textures from their low-resolution counterparts resulting in a low-fidelity images.
We employed a text-conditioned super-resolution model which is yet another diffusion model similar to the base text-to-image generator with Efficient U-Net as the image encoder-decoder module. The super-resolution model is conditioned on the text embeddings $e_T$ in addition to the low-resolution image produced by the previous diffusion model.
This integration of text and images allows for more targeted and context-aware generation of high-quality images, aligning with the provided textual guidance.

\section{Experiment and Results}
\label{sec:results}

\subsection{Experimental Setup}
The ${321.9}$M parameters Surgical Imagen is trained for 300K iterations with a batch size of $16$, using data parallelism to leverage $4 \times$V100 GPUs (32GB each). Our training set included approximately 101K image-text pairs from the CholecT50 dataset \cite{nwoye2022rendezvous}. The initial output dimensions were $64 \times 64 \times 3$, with super-resolution scaling by a factor of 4. We used Adafactor for the base model and Adam for the super-resolution model as in \cite{saharia2022photorealistic}. Inference involved 1000 diffusion steps to generate the final high-quality images.

\subsection{Evaluation Setup}
We evaluated our model using both human expert and automated methods. 
For automated evaluation, we used the Frechet Inception Distance (FID) score \cite{NIPS2017_8a1d6947} to measure image fidelity and the Contrastive Language-Image Pre-Training (CLIP) score \cite{radford2021learning} to evaluate alignment between the generated images and the input prompts. Additionally, we employed surgically meaningful metrics from \cite{lee2023holistic} to holistically and qualitatively assess the generated images for accuracy, reasoning, and knowledge revealing the latent trait of the model.

\begin{figure*}[!t]
    \centering
    \includegraphics[width=0.9\textwidth]{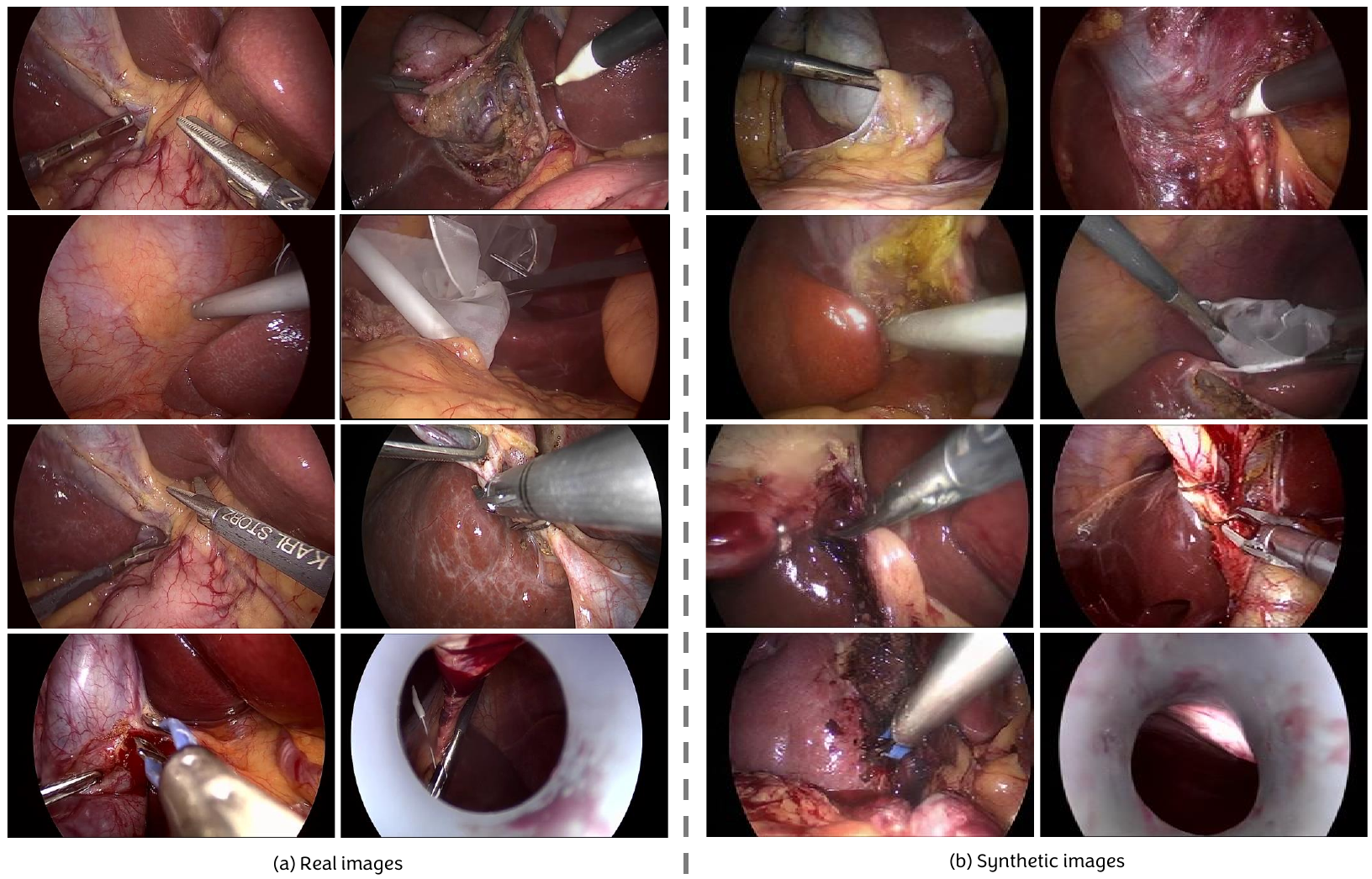}
    \caption{The quality of the (b) Surgical Imagen generated images in comparison with the (a) real surgical images of similar context.}
    \label{fig:realfake}
\end{figure*}

For human evaluation, we design a study in form of an online survey polling surgeons and medical professionals, with varying degree of expertise and levels of experience in laparoscopic cholecystectomy, to assessed the photorealism and alignment of the generated images with the text prompts.
The survey comprised 140 questions, each accompanied by four image options, totaling 640 images — approximately evenly split between real and generated. The survey was divided into two phases, taking around 9 minutes to complete. 
In the first phase, evaluators were asked to identify real images from a mix of actual and generated images across 6 randomly selected questions. In the second phase, consisting of 10 randomly selected questions, they were tasked with selecting images that best matched given prompts, focusing on assessing the alignment of tools, actions of tools, anatomies, and tool-tissue interactions.
Examples of survey questions include: "\textit{- Which of these images represents a coagulation of the liver? - Select the images corresponding to a grasper retracting the gallbladder. - Select the images where a tool is used to dissect the cystic plate. - Choose only the real images from options below.}
We analyzed the percentage of real and generated images identified as photorealistic and correctly aligned by respondents to gauge the model's performance effectively.
The survey and the responses can be accessed via the project page.
Additional details of the survey participants, including their demographics and expertise, are provided in the appendix.

\subsection{Results}
Some samples of the real vs generated images are shown in Fig. \ref{fig:realfake}. We evaluate the results in the subsequent sections.

\smallskip

\subsubsection{Quality}
Quality evaluates the photo-realism or visual realism of the generated surgical images to ensure they closely resemble authentic images, crucial for applications requiring accurate and convincing representations of surgical scenarios. Here, we employ both human expert and automated evaluation.
For the human evaluation, an initial survey with 88 respondents is analyzed in Table \ref{tab:survey}. Surgical Imagen achieves $37.6\%$ preference rate indicating high image quality generation compared to $64.3\%$ preference to the actual images. This shows how closely the generated images are to the real image such that  $\sim 38\%$ of the time, {experts fail} to identify the real images and $37.6\%$ of the time, {they classify} the synthetic images as real.
{A 37.6\% preference for generated images, while roughly half of the 64.3\% preference for real images, is highly significant in a field demanding extreme precision, such as surgery, where even subtle details like vein patterns are critical. This result underscores the realism of our generated images, as they are able to confuse expert surgeons nearly half the time, a remarkable achievement given the stringent visual requirements of the domain.}
For the automated evaluation, we report an FID score of 3.7 for Surgical Imagen, in Table \ref{tab:automated}, indicating closer similarity to real {images compared to the baseline method}.

\begin{table}[!t]
    \centering
    \caption{Result of human evaluation survey comparing real and generated images on photorealism and text-image alignment. }
    \label{tab:survey}
    \setlength{\tabcolsep}{23pt}
    \resizebox{0.99\linewidth}{!}{%
    \begin{tabular}{@{}lcc@{}}
        \toprule
         Images & Photorealism $\uparrow$ & Alignment $\uparrow$ \\
         \midrule
         Real & $64.3 \pm 0.2$ & $72.0 \pm 0.2$ \\
         Surgical Imagen & $37.6 \pm 0.1$ & $50.8 \pm 0.2$ \\
         \bottomrule
    \end{tabular}
    }
\end{table}

\begin{table}[!t]
    \centering
    \caption{Automated evaluation of real and generated images for fidelity (FID) and alignment (CLIP) scores.}
    \label{tab:automated}
    \setlength{\tabcolsep}{23pt}
    \resizebox{0.99\linewidth}{!}{%
    \begin{tabular}{@{}lcc@{}}
        \toprule
         Images & FID score $\downarrow$ & CLIP score $\downarrow$ \\
         \midrule
         Real & - - - & $23.01\pm0.1$ \\
         StackGAN \cite{zhang2017stackgan} & $5.83$ & $28.53\pm 0.9$ \\
         Surgical Imagen & $3.70$ & $26.84\pm 0.5$ \\
         \bottomrule
    \end{tabular}
     }
\end{table}

\begin{figure*}[!t]
    \centering
    \includegraphics[width=0.9\textwidth]{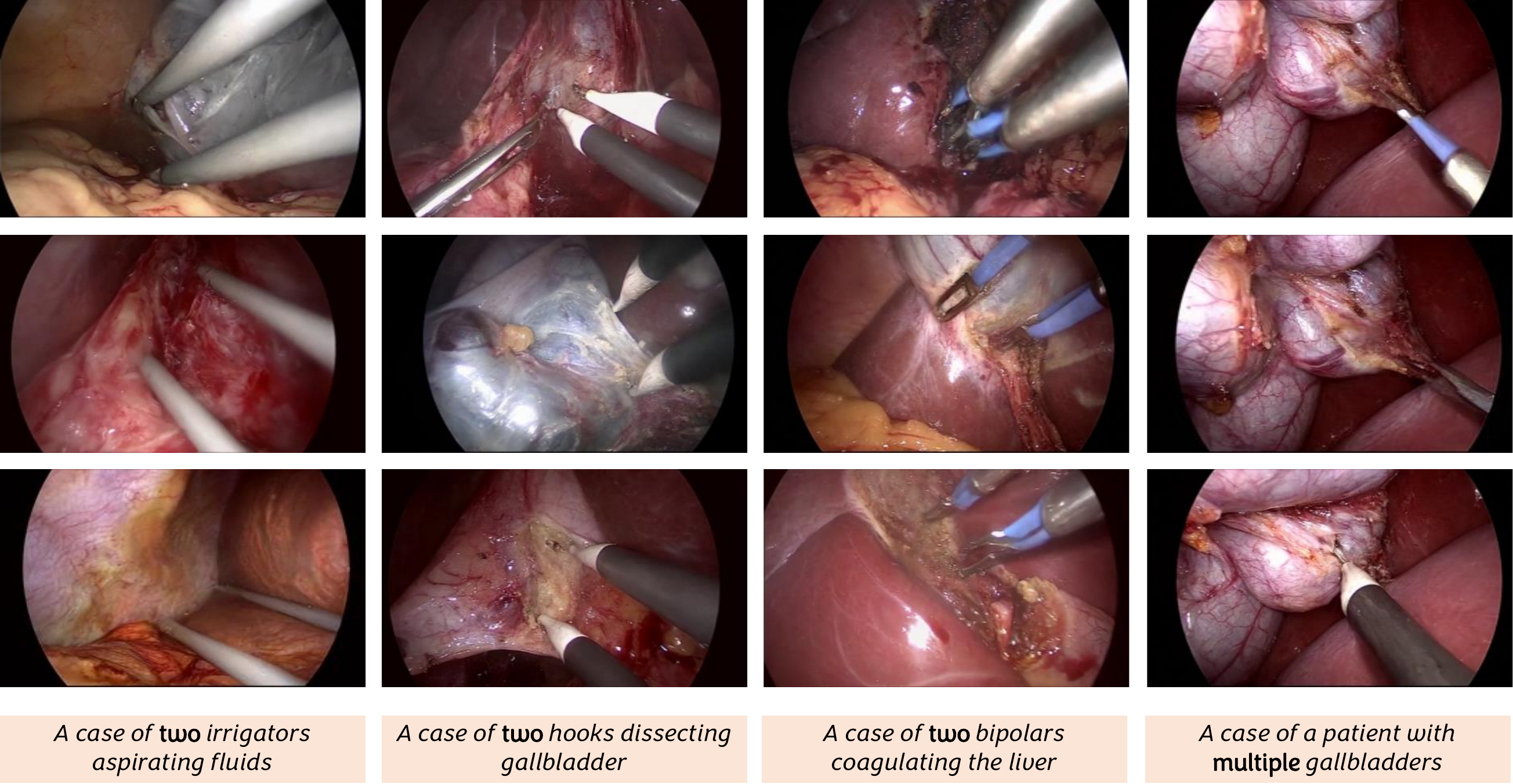}
    \caption{Generation of clinically impossible situations.}
    \label{fig:impossible}
\end{figure*}

\smallskip

\subsubsection{Task alignment}
This assessment is aim at verifying the semantic coherence between the generated images and the accompanying input text prompts, ensuring that the visual content accurately corresponds to the intended surgical context for effective communication and understanding. We also employ both human expert and automated evaluations to measure the degree of text-image alignment in the generated images.
For the expert evaluation, we measure how often the clinicians think that the AI generated images are aligned with the input text semantics. As reported in Table \ref{tab:survey}, the clinicians agree with Surgical Imagen $50.8\%$ of the time, meaning that the rate of disagreement is even less ($49.2$) suggesting our model’s ability to generate images that align well with real-world surgical activity based on three words captions.
On the other hand, the clinicians rate the real images alignment as correct $72.0\%$ of the time, leaving room for doubt ($28.0\%$) in such complex domain as surgery.
A 72\% alignment correctness score on real images reflects the meticulous nature and strict precision requirements of surgical action triplet recognition, where even experts often face challenges due to ambiguous anatomical boundaries and lack of temporal context. Achieving above 50\% alignment correctness on generated images under these stringent conditions highlights the ability of our model to produce data that aligns closely with expert expectations, demonstrating its potential to support complex surgical tasks.
Automated evaluation in Table~\ref{tab:automated} shows a CLIP score of $26.84$ for Surgical Imagen, quite similar to the $23.01$ for real images computed on a sample size of $10K$.

\smallskip
\noindent
Overall, the medical participants were highly challenged by the realistic characteristics of the samples, which validates their authenticity. 
The survey is left open for more responses from the global endoscopist community.

\begin{figure}[t!]
    \centering
    \includegraphics[width=1\linewidth]{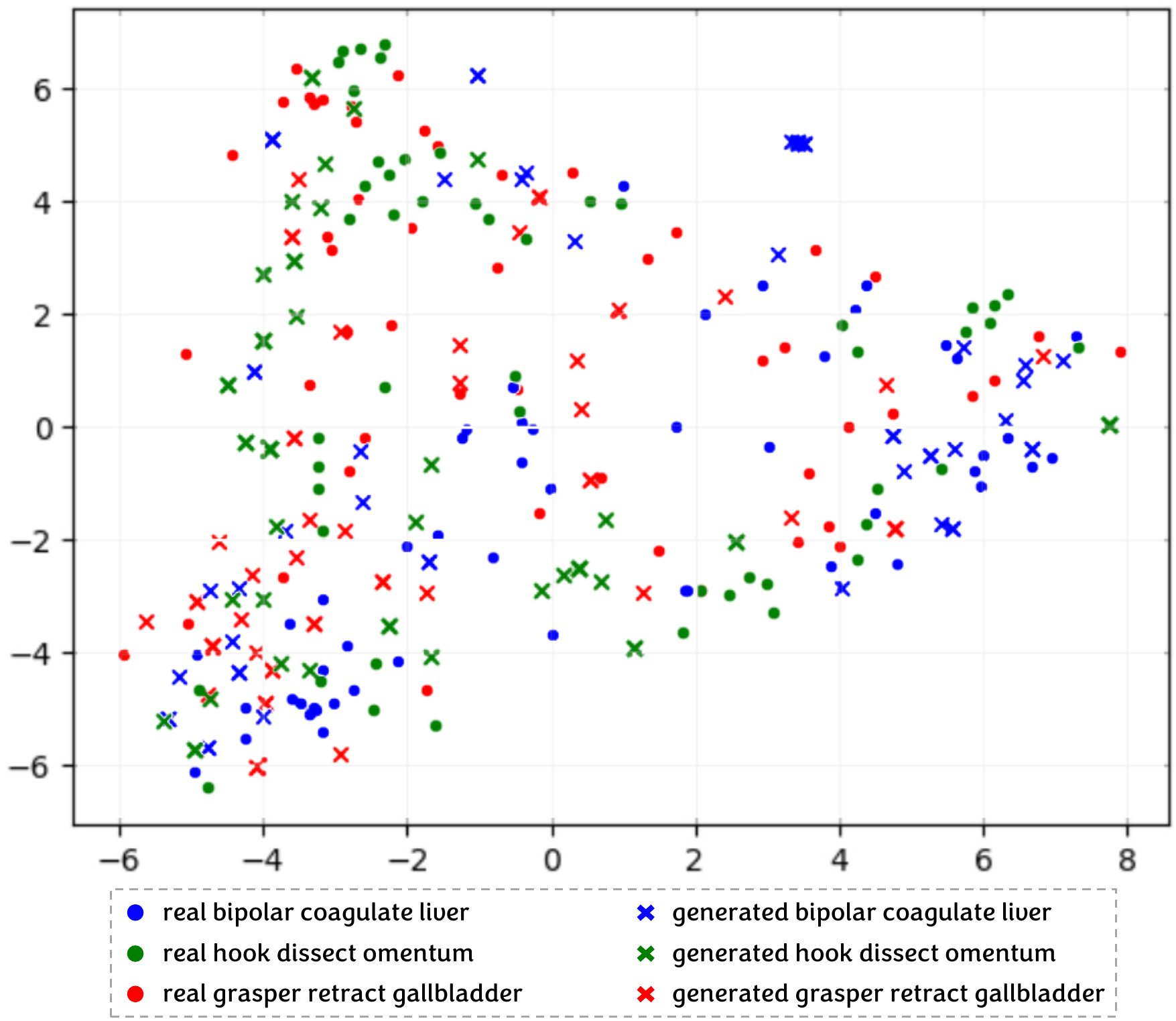}
    \caption{Feature proximity of selected samples of real and generated images.}
    \label{fig:proximity}
\end{figure}

\smallskip

\subsubsection{Reasoning}
Our interest here is to measure the compositionality in the generated images. We evaluate the model's ability to comprehend and correctly represent objects, counts, and spatial relationships within the generated surgical images, reflecting a deeper understanding of the underlying surgical processes and components. 
As shown in Fig. \ref{fig:impossible}, we explore the model into generating images where we encode the number of objects and their spatial relationship in the text. Even though surgical world is a control environment where the number of objects (tools, anatomy, etc) are fixed and known, we stretch the model into generating impossible situations such as patients with two gallbladders, the use of multiple instances of some instruments such as bipolar, clipper, hook, etc. which are never present in the training data.

\begin{figure*}[!t]
    \centering
    \includegraphics[width=0.9\textwidth]{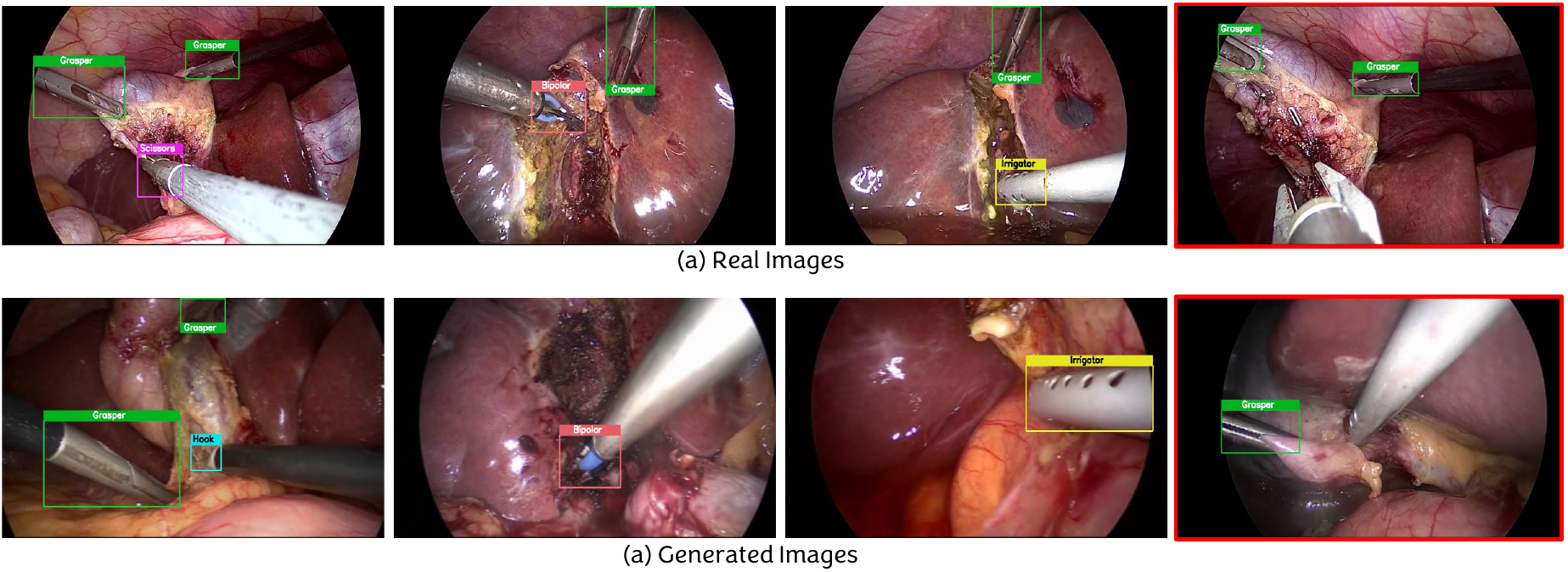}
    \caption{Tool detection results in the form of bounding box and class labels, inference by a pretrained SurgiTrack \cite{nwoye2024surgitrack}.}
    \label{fig:det}
\end{figure*}

\begin{figure*}[!t]
    \centering
    \includegraphics[width=0.9\textwidth]{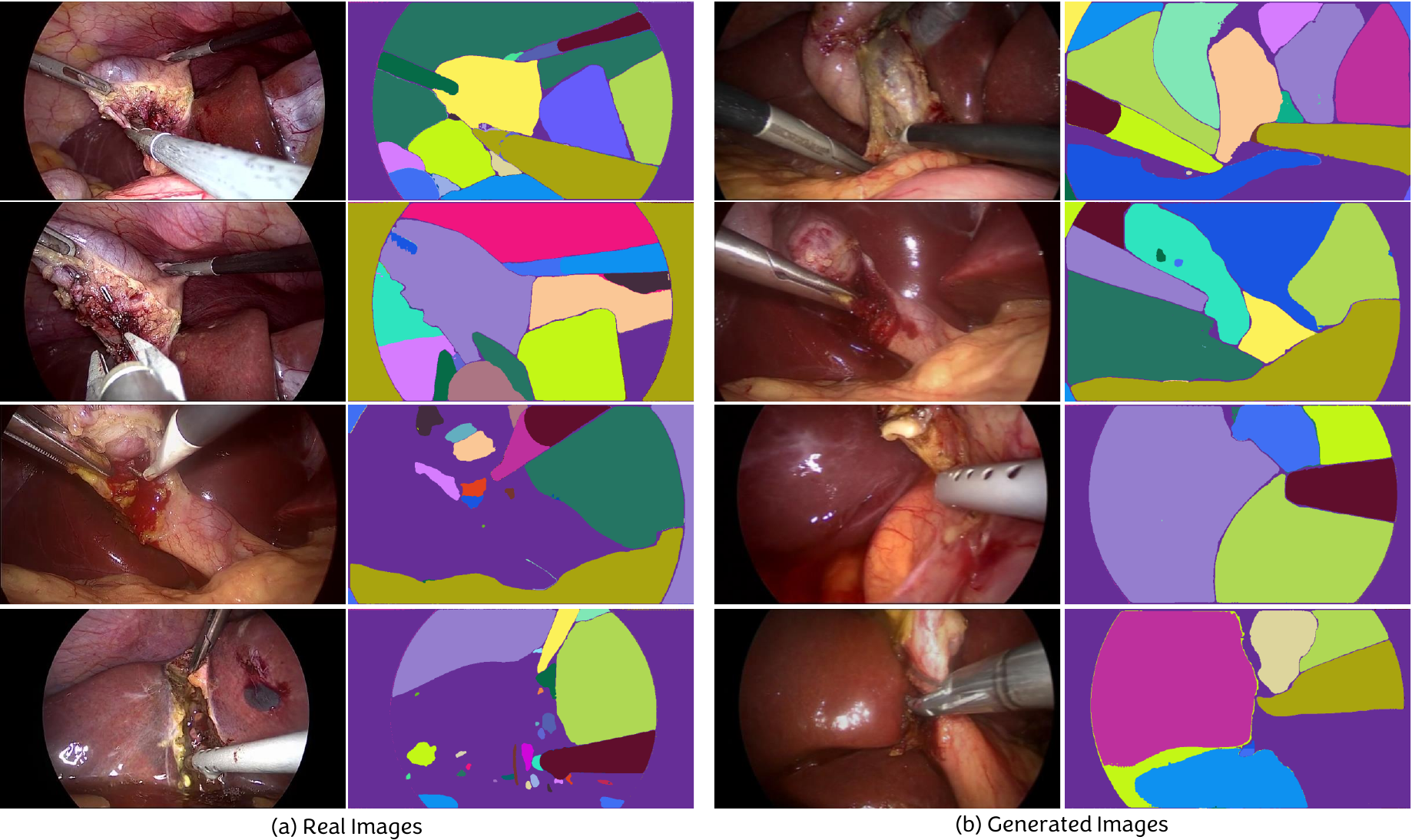}
    \caption{The results of unsupervised segmentation using SAM \cite{kirillov2023segment} on (a) real images and (b) generated images.}
    \label{fig:segm}
\end{figure*}

Apart from showing that the model is able to capture the real world reasoning in the synthetic data, it is important to show that it can convey that information, meaning that AI models should be able to learn from generated images and improve their performance. 
To ascertain this, we examine the feature proximity of generated and real images in the latent space, as depicted in Fig.~\ref{fig:proximity}.
While we observe some alignment of features, our current training of AI models on generated data has not shown substantial improvement. This discrepancy may stem from training objectives of the generative model primarily focused on achieving photorealism and visual content alignment rather than ensuring perceptual alignment of latent features with real data. Future research will delve deeper into this aspect to enhance the model's ability to facilitate learning and performance improvement from generated images.


\begin{table}[!t]
    \centering    
    \caption{Inference results (\%) on real vs generated data using pretrained deep learning models. 
    }
    \label{tab:classifier}
    \setlength{\tabcolsep}{6pt}
    \resizebox{0.99\linewidth}{!}{%
    \begin{tabular}{@{}lcclcc@{}}
        \toprule
        \multirow{2}{*}{Task [Method]} & \multicolumn{2}{c}{Real Dataset} & \phantom{abc} & \multicolumn{2}{c}{Generated Data }\\ \cmidrule{2-3} \cmidrule{5-6}
        & Acc $\uparrow$ & AP $\uparrow$ && Acc $\uparrow$ & AP $\uparrow$ \\
         \midrule
         Tool recognition \cite{nwoye2019weakly} & 77.3 & 93.0 && 77.9 & 92.7 \\
         Triplet recognition \cite{nwoye2022rendezvous}  & 26.3 & 23.2 && 16.8 & 13.9\\
         \bottomrule
    \end{tabular}
    }
\end{table}

\smallskip

\subsubsection{Knowledge}

The knowledge metric assesses if the model have knowledge about the world or domains. The essence is to gauge the model's awareness of real-world surgical domains, ensuring that the generated images reflect accurate and contextually relevant information, enhancing the educational and practical value of the synthetic images.
Here, we rely on tool classifier (results in Table \ref{tab:classifier}), tool detectors (results in Fig. \ref{fig:det}) and instance segmentation (results in Fig. \ref{fig:segm}) to evaluate if the generated images contain domain knowledge identifiable by AI models that are trained on the real images.

The results shows the tool classification accuracy and mean average precision similar for both real and generated images (see Table \ref{tab:classifier}). However, there is a $10\%$ difference on triplet recognition scores. 
For lack of groundtruth labels, we evaluate the detection and segmentation task qualitatively only.
The detection results using SurgiTrack \cite{nwoye2024surgitrack} localize and correctly classify the tools in both real and generated images. Similarly, SAM \cite{kirillov2023segment} segments the tools and great portion of the anatomies in both real and generated images.

\smallskip

\subsubsection{Impact of Instrument-based Class Balancing}
\begin{figure}[!t]
    \centering
    \includegraphics[width=0.9\linewidth]{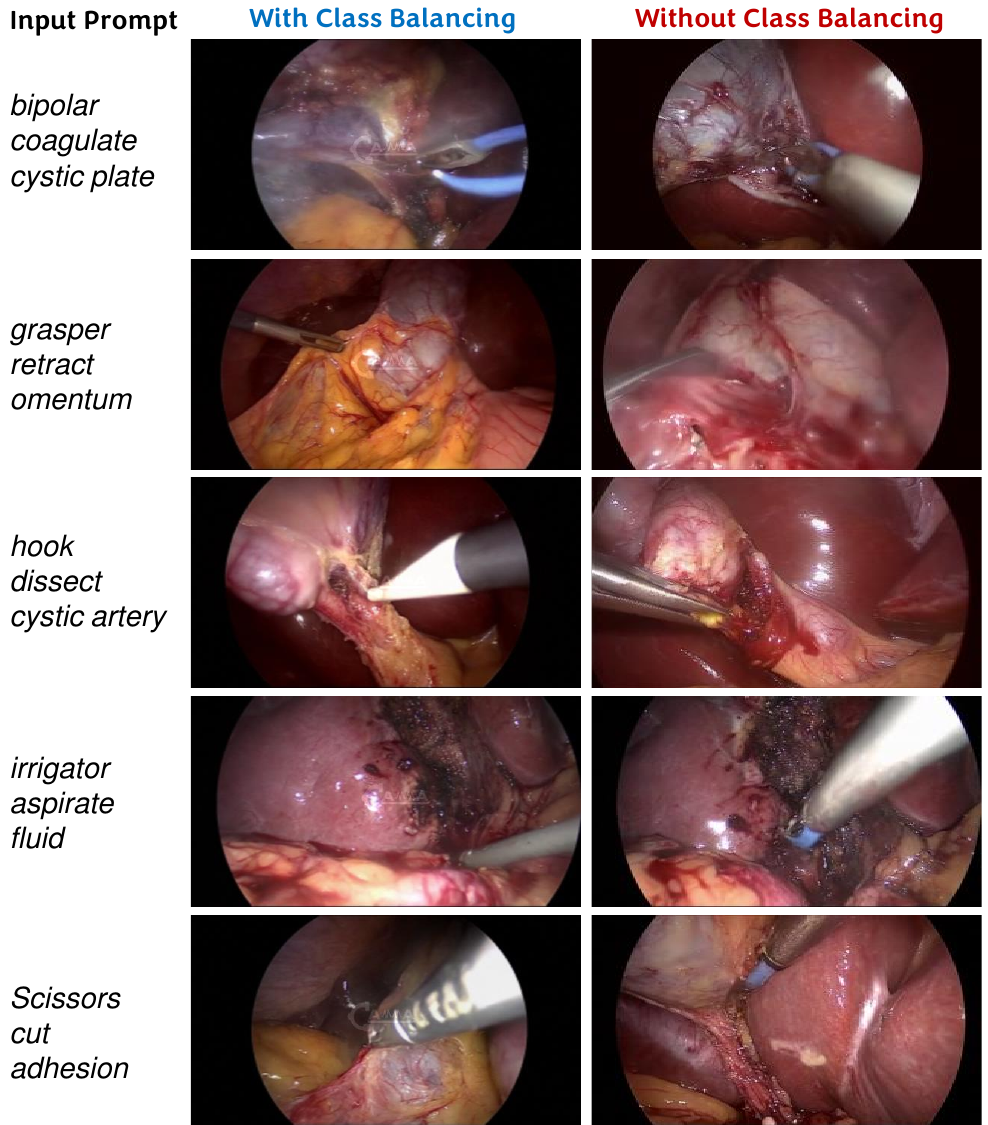}
    \caption{Sample output of Surgical Imagen with and without instrument-based class balancing technique.}
    \label{fig:ablation-cb}
\end{figure}

In the ablation study, we compare the outputs of the preliminary model, which lacked the instrument-based class balancing technique, with those of the improved model that incorporates this technique. Without class balancing, the preliminary model showed a bias towards generating images corresponding to the most frequent triplet classes, such as ``\textit{grasper retract gallbladder}" or ``\textit{bipolar coagulate gallbladder}", regardless of the input prompts. This resulted in limited diversity and accuracy in the generated images. In contrast, the model with class balancing demonstrated improved convergence, producing a more diverse set of images that aligned better with the input text prompts. 
We demonstrate this with less frequent surgical action triplets in Fig. \ref{fig:ablation-cb}. Comparison with these preliminary results provide empirical evidence that the instrument-based class balancing technique mitigates data imbalance and enhances the model’s ability to generate semantically accurate and diverse outputs.

\smallskip

\subsubsection{Ablation Study on Super-resolution}
\begin{figure*}
    \centering
    \includegraphics[width=0.9\textwidth]{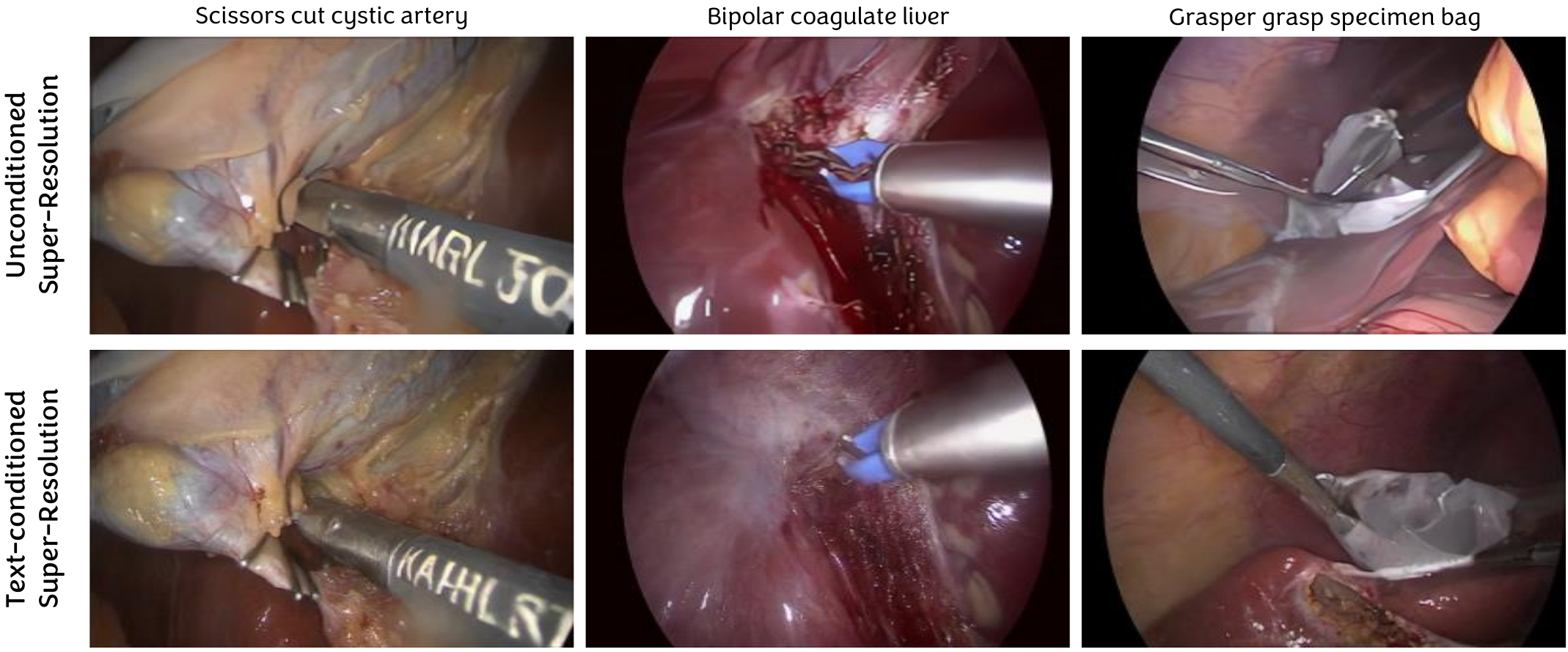}
    \caption{The quality of the synthetic images in unconditioned vs text-conditioned super-resolution model. The unconditioned super-resolution model upsamples low resolution images by pixel interpolation, whereas the text-conditioned counterpart upscales with semantic details including finer anatomical structures and smokes from coagulation device.}
    \label{fig:super}
\end{figure*}

We show in Fig. \ref{fig:super} the distinction between using an unconditioned and text-conditioned super-resolution models. The text-conditioned super-resolution model shows prowess in retaining the fine-grained details in the anatomical structures and generating more photo-realistic images than the unconditioned counterpart. Upon closer visual inspection, it is evident that the unconditional super-resolution model enhances the image uniformly, thus lacking the capability to focus on the actions being performed. Using triplet based captions as condition improves the generation capability by adding crucial details in the zones mentioned in the text, hence generating more realistic images.

\section{Discussion}
\label{sec:conclusion}

The introduction of \textit{Surgical Imagen} represents a significant leap forward in the field of surgical AI, leveraging the T5 language model and the CholecT50 dataset to generate highly realistic surgical images from concise triplet-based textual prompts. This approach not only addresses the longstanding challenges associated with acquiring annotated surgical data but also sets a precedent for enhancing the efficiency and scope of surgical AI applications.

Central to our findings is the validation of triplet-based labels as effective descriptors for capturing intricate surgical actions. This discovery carries profound implications for surgical AI research, suggesting that succinct and semantically robust annotations can streamline model development and improve task specificity. Moreover, our methodological innovation in addressing class imbalances within the dataset demonstrates a crucial advancement in ensuring the robustness and applicability of AI models trained on medical data.

The comprehensive evaluation of \textit{Surgical Imagen} using both human expert assessments and automated metrics underscores its capability to produce images that are not only visually accurate but also contextually aligned with textual prompts. These results affirm the model’s reliability across various evaluation criteria and highlight its potential to fostering research and clinical applications.

\subsection{Potential Clinical Applications}
The clinical use of this model extends across various domains, encompassing educational applications, content generation, and simulation.

\smallskip

\subsubsection{Simulation}
Simulation stands as the primary domain in which we anticipate the widespread adoption of Surgical Imagen. A pivotal forthcoming advancement of the model would empower us to generate complete videos, thus sparking discourse within the simulation community. These materials are poised to play a pivotal role in broadening access to an extensive array of synthetic surgical laparoscopic procedures, thereby democratizing their availability. The capability to produce images affords us the opportunity to replicate frames depicting laparoscopic complications, a facet often underrepresented in existing datasets due to data scarcity. Beyond addressing complications, our system excels in manipulating camera angles, thereby reproducing sub-optimal views: a feature that has already demonstrated its efficacy in reducing the learning curve within surgical training environments \cite{RN1}. 
Furthermore, the application of Surgical Imagen extends to virtual reality (VR). This entails training and refining the model using publicly accessible stereo view datasets sourced from robotic surgeries (e.g. DaVinci), with the objective of generating paired images. These images could subsequently be integrated into VR headsets for immersive 3D simulation, promising enhanced training efficacy characterized by reduced time investment and errors \cite{RN2}.

\smallskip

\subsubsection{Education}
The presence of a model capable of comprehensively grasping and accurately replicating the surgical setting will influence education across various dimensions. As previously mentioned, the first dimension is simulation. The second, which is more intrinsic to educational methodologies, draws parallels with the impact of 3D virtual anatomical atlases \cite{RN3}. The ability to deconstruct a surgical procedure into its phases, coupled with a detailed description of actions using the triplet formulation, and the capacity to visualize various iterations, provides students and residents with a straightforward means to thoroughly explore the intricacies of a procedure.

\smallskip

\subsubsection{Content generation}
As outlined earlier, content generation tailored for simulation purposes significantly influences the field. However, beyond simulation, there are additional use cases that stand to benefit. Among these, the production of clear and concise content effectively supports clinical presentations and the drafting of proposals. These same images could also facilitate communicating the procedural steps to patients, enhancing their understanding of surgical processes. Importantly, it's worth noting that a key advantage shared among all these applications is the absence of privacy concerns.

\smallskip

\subsubsection{Beyond Laparoscopy}
Surgical Imagen lays the foundation for an ambitious undertaking that is both unprecedented and challenging. This initiative enables the seamless integration of diverse modalities. As an example, an ultrasound image could be transformed into textual information and subsequently utilized by Imagen to produce a synthetic laparoscopic image. Although this prospective application poses certain technical challenges that require resolution, we are confident that such a model will expedite research endeavors in this domain. 

\subsection{Limitation and Societal Impacts}

Despite its strengths, \textit{Surgical Imagen} has limitations that warrant consideration. The reliance on the CholecT50 dataset, while rich, may not fully represent the diversity of surgical procedures and contexts, potentially limiting the generalizability of our findings. Future research should explore additional datasets and surgical domains to further validate and refine our model.

Computational resources pose another challenge, as the diffusion-based model requires significant compute time to generate images. While increasing GPU resources could mitigate this issue, the high demand for GPUs can still be a bottleneck, potentially limiting accessibility for smaller research teams and institutions. Implementing optimizations and leveraging more efficient hardware present alternatives that could make the model more practical for widespread use.

Additionally, the content produced by \textit{Surgical Imagen} is subject to errors, necessitating continuous assessment of clinical validity. Similar to our first survey, regular validation by clinical experts is essential to ensure the accuracy and reliability of the generated images. These evaluations should be frequent to allow the model to enter an active learning loop, where feedback from clinicians can be used to improve the generation process over time.

Ethical considerations are also paramount, particularly regarding the use of synthetic data in medical training and decision-making. It is crucial to maintain transparency in the development and application of such models to build and sustain trust in clinical practice. Ethical guidelines should be established to govern the use of synthetic data, ensuring it supplements rather than replaces real-world data.

The possible societal impacts of \textit{Surgical Imagen} include the potential for significant improvements in surgical education and patient safety. By providing realistic and varied surgical scenarios, the model can enhance training and preparedness among medical professionals. However, there is a risk of over-reliance on synthetic data, which could lead to complacency or inaccuracies in real-world applications. Continuous scrutiny, rigorous validation, and a balanced integration of synthetic and real data are imperative to mitigate these risks and maximize the benefits of this innovative technology.

\subsection{Future direction}
Future work will focus on refining the model to handle more complex surgical scenes, involving dynamic interactions and multiple tools. Expanding to surgical video generation, which incorporates temporal data, will improve realism and comprehensiveness. It can also be extended to 3D volumetric data, such as CT/MRI scans or 3D ultrasound, to generate synthetic datasets for preoperative planning and education. It could also leverage depth-enhanced laparoscopic video or 3D organ reconstructions for training on spatially complex surgical scenarios. Integrating real-time feedback from surgical practitioners will further enhance model accuracy, ensuring alignment with current clinical practices and evolving techniques, ultimately supporting better surgical training and decision-making.

\section{Conclusion}
In this work, we introduced \textit{Surgical Imagen}, a diffusion-based text-to-image generative model designed to address the challenges of obtaining high-quality surgical data for research and education. Utilizing the CholecT50 dataset, which provides surgical images annotated with triplet-based labels (instrument, verb, target), we conducted a thorough analysis to determine the most suitable language model for obtaining representative text embeddings, finding T5 to be superior. Our model, built on the Imagen framework, integrates an innovative instrument-based class balancing technique to counteract dataset imbalances and improve convergence. Evaluated using both human expert surveys and automated metrics like FID and CLIP scores, Surgical Imagen demonstrates its ability to generate photorealistic, contextually accurate surgical images from simple textual prompts. Our results underscore the potential of generative models in creating synthetic surgical datasets, setting a new benchmark for future research in surgical AI. 
\section*{Acknowledgements}
This work was supported by French state funds managed within the Plan Investissements d’Avenir by the ANR under references ANR-20-CHIA-0029-01 (National AI Chair AI4ORSafety), ANR-10-IAHU-02 (IHU Strasbourg) and by BPI France (Project 5G-OR). This work has also received funding from the European Union (ERC, CompSURG, 101088553). Views and opinions expressed are however those of the authors only and do not necessarily reflect those of the European Union or the European Research Council. Neither the European Union nor the granting authority can be held responsible for them. 

This work was granted access to the servers/HPC resources managed by CAMMA, IHU Strasbourg, Unistra Mesocentre, and GENCI-IDRIS  [Grant 2021-AD011011638R3, 2021-AD011011638R4]. 

The authors acknowledge the following clinicians for their participation in filling the evaluation survey:
Adolfo Gutierrez,
Alain Garcia (IHU Strasbourg),
Alicia López (A Coruña University Hospital),
Amir Ashraf Ganjouei (UCSF),
Ana (Otero),
Andrea Balla (University Hospital Virgen Macarena),
Andrea Rosati (Fondazione  Policlinico Gemelli, Rome),
Anna Minasyan (Coruña University Hospital, Coruña, Spain),
AS Soares (Hospital prof. Dr. Fernando Fonseca, UCL),
Clarisa (Ponderas Academic Hospital),
Donald Van Der Fraenen (Maria middelares Gent),
Federica (University of Turin),
Felipe Asenjo de Castro (Hospital Central de la Defensa “Gomez Ulla” Madrid)
Florent Alexandre (Strasbourg IHU),
Francesco Marzola (University of Turin - MITIC),
Gabriel Szydlo Shein (Université de Strasbourg),
Georgios Kourounis (Newcastle University),
Giovanni Guglielmo Laracca (Sant’Andrea Hospital, Sapienza University of Rome, Italy),
Giuseppe Massimiani (Università Cattolica del Sacro Cuore),
Helmuth Radrich (Karl Storz VentureONE),
Joris Vangeneugden (Ghent University Hospital),
Juan Manuel Verde (IHU Strasbourg),
Julieta Montanelli (IHU Strasbourg),
Lachlan Dick (University of Edinburgh),
Leonardo Sosa Valencia,
Luca Gordini (ASL Sulcis),
Luigi Boni (Fondazione IRCCS Ca' Granda Ospedale Maggiore Policlinico di Milano, University of Milan),
Manish Jethani (Kokilaben Dhirubhai Ambani Hospital Mumbai India),
Marta Goglia (Sapienza University of Rome),
Matteo Pavone (Policlinico Gemelli/IHU Starsbourg/IRCAD Strasbourg),
Michele Ammendola ('Magna Graecia' University \& 'Renato Dulbecco' University Hospital, Catanzaro, Italy),
Muhammad Umar Younis (Mediclinic City Hospital Dubai),
Nabani Banik (Université de Strasbourg),
Patricia Deltell (Hospital general Alicante),
Pietro Leoncini (University of Turin MITIC),
Po-Hsing Chiang (Chang Gung Memorial Hospital, Taiwan),
Roberto Spagnulo (MITIC- UniTO),
Romina (Braverman),
Salvador Morales-Conde (University Hospital Virgen Macarena, Sevilla, Spain),
Sofia Teixeira da Cunha (Kantonsspital Olten),
Sonia Pérez-Bertólez (Hospital Sant Joan de Déu),
Yanhua Zhang (PoliTo) and others (please see full list: \url{https://camma-public.github.io/endogen/surgical-imagen/index.html#thanks}).

\section*{Conflict of Interest}
The authors has no conflict of interest to declare.



\onecolumn
\appendix

\subsection{Survey design}

\begin{figure}[!h]
    \centering
    \includegraphics[width=\linewidth]{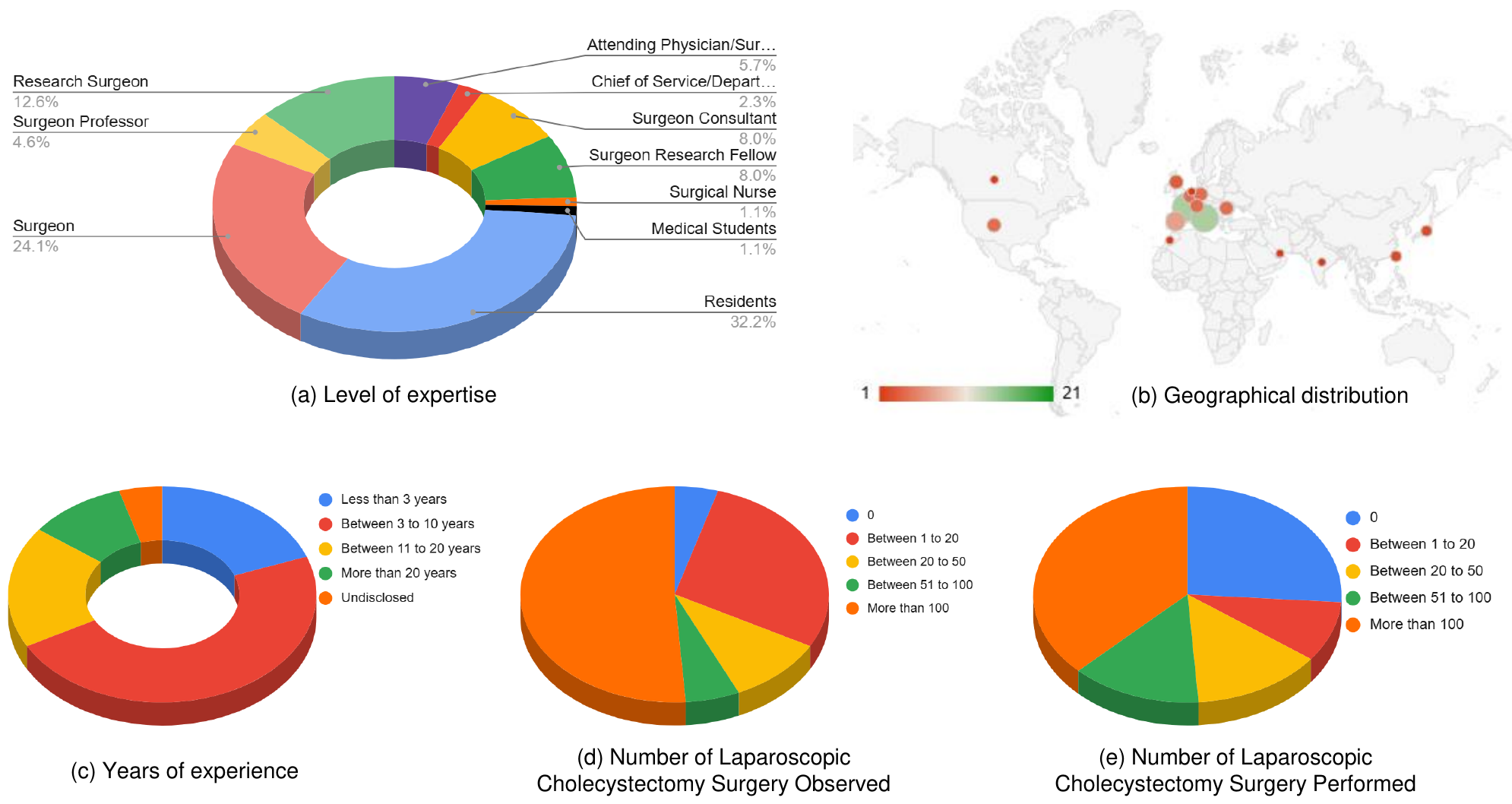}
    \caption{Survey design: participant demographics and professional backgrounds. The figure provides a breakdown of participants' expertise levels, years of surgical experience, the number of laparoscopic cholecystectomies observed and performed, and the geographic distribution of their affiliations, offering a comprehensive overview of the survey cohort.}
    \label{fig:enter-label}
\end{figure}

\end{document}